\documentclass[11pt]{article}

\usepackage[margin=1in]{geometry}
\usepackage{amsmath,amssymb,mathtools}
\usepackage{booktabs}
\usepackage{array}
\usepackage{enumitem}
\usepackage{microtype}
\usepackage{graphicx}
\usepackage{url}
\usepackage[hidelinks]{hyperref}
\usepackage[round,authoryear]{natbib}
\usepackage{xcolor}
\usepackage{xspace}

\newcommand{\argmin}{\operatorname*{arg\,min}}
\newcommand{\rc}{\textsc{Rollcast}\xspace}

\title{\textbf{Rollcast: Proper-Score Gated Rolling Anchors for Adaptive Probabilistic Time-Series Forecasting}}
\author{Giancarlo Vercellino\\
Independent Researcher\\
\texttt{giancarlo.vercellino@gmail.com}}
\date{September 2026}

\begin{document}
\maketitle

\begin{abstract}
We introduce \rc, a probabilistic forecasting method for a single numeric time series that combines simple rolling statistics through a state-dependent predictive mixture. Unlike conventional forecast combinations whose components are separately fitted models, \rc uses a compact dictionary of rolling \emph{anchors}: the local mean, median, minimum, maximum, a one-step linear-regression endpoint, and selected empirical quantiles. Historical forecast origins are represented by the anchors' robustly scaled displacement from the current observation. Each anchor is paired with a causal, state-conditional residual distribution estimated from similar earlier states, while a softmax gate is fitted by minimizing regularized negative log predictive density of the full mixture. Gate probabilities are stabilized by an adaptive persistence rule and multi-step distributions are generated recursively through an all-anchor particle mixture.

The CRAN implementation is evaluated in a synthetic calibration benchmark with eight pre-specified data-generating processes, 250 independent replications per process, 300 training observations, and forecast horizons one through six. Compared with the true-DGP oracle, mean coverage is 0.862 versus 0.901 for nominal 90\% intervals and 0.915 versus 0.952 for nominal 95\% intervals. Rollcast intervals are nevertheless 13.6\% and 15.9\% wider than the oracle at those levels, while normalized CRPS is 14.4\% above the oracle overall. Coverage declines with horizon, from 0.898 to 0.840 at the 90\% level. Performance is closest to the oracle under stochastic volatility, heavy-tailed autoregression, Gaussian autoregression, threshold dynamics, and variance breaks; local trend and Markov-switching dynamics are more difficult. These results locate the main weakness in multi-step conditional calibration rather than interval width alone.
\end{abstract}

\textbf{Primary arXiv category:} stat.ME (Statistics Methodology).\\

\section{Introduction}
Forecast combination has a long history in time-series analysis, from the original evidence that weighted combinations can outperform individual forecasts \citep{BatesGranger1969} through broad reviews of why simple and adaptive combinations can be effective \citep{Clemen1989,WangEtAl2023}. The literature now includes predictive-density pooling, time-varying weights, feature-dependent combinations, and mixture-of-experts architectures \citep{HallMitchell2007,GewekeAmisano2011,LiKangLi2023,NiEtAl2024}. In parallel, probabilistic forecasting has emphasized that the complete predictive distribution should be evaluated with proper scoring rules rather than only point accuracy \citep{GneitingRaftery2007,GneitingKatzfuss2014}.

\rc starts from a deliberately different object than most forecast-combination methods. Its components are not separately estimated ARIMA, state-space, regression, or machine-learning models. They are simple summaries of a recent rolling window: a mean, median, extrema, selected quantiles, and a short linear-regression endpoint. These summaries are treated as low-capacity \emph{forecast anchors}. Their purpose is not to be individually correct, but to provide a small set of interpretable local hypotheses whose relevance can change with the observed state.

Two statistical traditions are especially close to this construction. Local analog and nearest-state forecasting uses historical states similar to the current one to infer subsequent behavior \citep{FarmerSidorowich1987,SugiharaMay1990}, while nonparametric conditional-density estimation uses local weighting and kernels to estimate a full response distribution conditional on covariates \citep{HyndmanEtAl1996,HyndmanYao2002}. \rc uses nearest historical states to build anchor-specific residual distributions and combines those distributions with a state-dependent gate trained by a proper score.

The resulting method has four operational layers. First, the current anchor configuration is expressed relative to the latest level and a robust local scale. Second, each anchor receives a residual distribution estimated from causally available historical states that resemble the current state. Third, a softmax gate is trained on the logarithmic score of the complete predictive mixture, aligning weight learning with density quality. Fourth, the one-step mixture is propagated recursively, with gate inertia adapting to the amount of state movement.

Relative to these literatures, \rc organizes proper-score pooling, nearest-state retrieval, kernel residual smoothing, and dynamic mixture weights around a compact dictionary of rolling statistical anchors. Anchor geometry supplies the gating state; residual distributions are estimated around every anchor using responsibility-weighted neighbors; gate persistence depends on observed state movement; and tuning uses common causal validation origins.

\section{Forecast anchors and state representation}
Let $y_1,\ldots,y_n$ be a univariate time series. At forecast origin $t$, define a rolling window of width $W$,
\begin{equation}
\mathcal{W}_t=(y_{t-W+1},\ldots,y_t).
\end{equation}
Let $A_{tj}$, $j=1,\ldots,M$, denote $M$ anchor functionals of this window. The default dictionary contains
\begin{equation}
\left\{\bar y_t,\; \operatorname{med}_t,\; \min_t,\; \max_t,\; \widehat y^{\mathrm{reg}}_{t+1},\; Q_t(q_1),\ldots,Q_t(q_Q)\right\},
\end{equation}
where $M=5+Q$ and the median is not duplicated among the user-specified quantiles.

The regression anchor is the ordinary least-squares fitted level one step beyond the rolling window. Writing $k=1,\ldots,W$, $\bar k=(W+1)/2$, and $\bar y_t=W^{-1}\sum_k y_{t-W+k}$,
\begin{align}
\widehat b_t &= \frac{\sum_{k=1}^W (k-\bar k)(y_{t-W+k}-\bar y_t)}{\sum_{k=1}^W(k-\bar k)^2},\\
\widehat y^{\mathrm{reg}}_{t+1} &= \bar y_t+\widehat b_t\{(W+1)-\bar k\}.
\end{align}
The one-step endpoint is used instead of the mean fitted regression value, because with an intercept the latter equals the sample mean.

Let $s_t>0$ be a robust local scale. The reference implementation uses the median absolute deviation multiplied by 1.4826, with IQR, standard deviation, range, and a small numerical floor as fallbacks for degenerate windows. The default relative state vector is
\begin{equation}
 x_{tj}=\frac{A_{tj}-y_t}{s_t},\qquad j=1,\ldots,M. \label{eq:relative-state}
\end{equation}
An optional raw-state formulation uses $x_{tj}=A_{tj}$. For fitting the gate and computing nearest states, $x_t$ is standardized componentwise using training-period means and standard deviations.

For each historical origin, define the standardized anchor error
\begin{equation}
 e_{tj}=\frac{y_{t+1}-A_{tj}}{s_t}, \label{eq:resid}
\end{equation}
and its absolute version $d_{tj}=|e_{tj}|$. Rather than assigning each observation exclusively to the ex post closest anchor, \rc uses a soft responsibility
\begin{equation}
 r_{tj}=\frac{\exp\{-(d_{tj}-\min_{\ell}d_{t\ell})/\tau\}}
 {\sum_{m=1}^{M}\exp\{-(d_{tm}-\min_{\ell}d_{t\ell})/\tau\}}, \label{eq:responsibility}
\end{equation}
where $\tau>0$ controls how sharply responsibility concentrates on the nearest anchor.

\section{Causal state-conditional component distributions}
A deterministic anchor alone does not provide a density forecast. \rc constructs an anchor-specific predictive component by borrowing residual behavior from similar earlier states. Let $z_t$ denote the standardized state corresponding to $x_t$. For a target training row $t$, only rows $r<t$ are eligible. Let $\mathcal{N}_t$ contain the $K$ nearest eligible states under root-mean-square Euclidean distance, and define state-similarity weights
\begin{equation}
\kappa_{tr}=\exp\left[-\frac{\|z_t-z_r\|_2^2/M}{2h_x^2}\right],\qquad r\in\mathcal{N}_t,
\end{equation}
with state bandwidth $h_x>0$.

Responsibilities are slightly smoothed to avoid zero support,
\begin{equation}
\widetilde r_{rj}=(1-\epsilon)r_{rj}+\frac{\epsilon}{M},
\end{equation}
where $\epsilon\in[0,1]$. The combined neighbor/anchor weight is $w_{trj}=\kappa_{tr}\widetilde r_{rj}$.

Let $\gamma\in[0,1]$ control the strength of residual correction. A continuous approximation to the component density at a realized one-step outcome can be written on the residual scale as
\begin{equation}
 \widehat f_{tj}(y_{t+1})=
 \frac{1}{s_t}
 \frac{\sum_{r\in\mathcal N_t} w_{trj}\,\phi_{h_{\mathrm{eff}}}
 \left(e_{tj}-\gamma e_{rj}\right)}
 {\sum_{r\in\mathcal N_t}w_{trj}}, \label{eq:component-density}
\end{equation}
where $\phi_h(u)=h^{-1}\phi(u/h)$ and
\begin{equation}
 h_{\mathrm{eff}}=\max(\gamma h_e,h_0).
\end{equation}
Here $h_e$ is the residual kernel bandwidth and $h_0>0$ is a \emph{validation-only} numerical floor. The floor keeps logarithmic scores finite at $\gamma=0$; it is not injected into actual forecast simulations. When $\gamma=0$, the deployed forecast is a pure discrete mixture over the anchors.

For simulation at a new state, a historical residual center $e_{rj}$ is sampled with probability proportional to $w_{trj}$, then Gaussian kernel jitter with standard deviation $h_e$ is added. Conditional on anchor $j$, a one-step candidate is
\begin{equation}
 Y^{(j)}_{t+1}=A_{tj}+\gamma s_t\widetilde e_{tj}. \label{eq:candidate}
\end{equation}
This separates \emph{where} an anchor places predictive mass from \emph{how much} local residual dispersion is admitted around it.

\section{Proper-score gating}
Let $\pi_{tj}$ denote the state-dependent probability assigned to anchor $j$. The gate is a multinomial logistic model on standardized states. Let $\widetilde z_t=(1,z_t^\top)^\top$ and let $\beta_j$ be the coefficient vector for component $j$, with one component fixed as the zero-reference vector for identifiability. Then
\begin{equation}
 \pi_{tj}=\frac{\exp(\widetilde z_t^\top\beta_j)}{\sum_{m=1}^{M}\exp(\widetilde z_t^\top\beta_m)}. \label{eq:gate}
\end{equation}

Crucially, the gate is not trained to reproduce the ex post label $\argmin_j d_{tj}$. Instead, it is fitted to maximize the probability assigned by the \emph{mixture density} to realized observations. For a training index set $\mathcal T$, the penalized negative log predictive density is
\begin{equation}
\mathcal L(B)=
-\frac{1}{|\mathcal T|}\sum_{t\in\mathcal T}
\log\left(\sum_{j=1}^{M}\pi_{tj}\widehat f_{tj}(y_{t+1})\right)
+\frac{\lambda}{2}\sum_{j=1}^{M-1}\|\beta_{j,-0}\|_2^2, \label{eq:loss}
\end{equation}
where the intercept is unpenalized and $\lambda\ge0$. This is directly aligned with the logarithmic proper score \citep{GneitingRaftery2007}.

The reference implementation optimizes this regularized objective numerically with BFGS. The important distinction is practical: the gate is rewarded for assigning probability to components that improve the density of the realized next observation, not simply for reproducing a hard ``best-anchor'' label.

\section{State-adaptive persistence}
A raw gate may change too quickly when neighboring states are nearly identical, or too slowly if persistence is imposed uniformly. \rc therefore makes inertia depend on standardized state movement. Let
\begin{equation}
 \delta_t=\left\{\frac{1}{M}\sum_{j=1}^{M}(z_{tj}-z_{t-1,j})^2\right\}^{1/2}.
\end{equation}
The persistence parameter is
\begin{equation}
 \rho_t=\rho_{\min}+(\rho_{\max}-\rho_{\min})\exp(-\delta_t/d_\rho), \label{eq:rho}
\end{equation}
with $0\le\rho_{\min}\le\rho_{\max}\le1$ and $d_\rho>0$. The stabilized gate is
\begin{equation}
 \widetilde\pi_t=\rho_t\widetilde\pi_{t-1}+(1-\rho_t)\pi_t. \label{eq:inertia}
\end{equation}
The rule has a direct interpretation. If the standardized state barely changes, $\rho_t$ stays close to $\rho_{\max}$ and the previous mixture receives more weight. As state movement grows, $\rho_t$ smoothly approaches $\rho_{\min}$ and the gate responds more quickly to the new state. The decay parameter $d_\rho$ controls how rapidly this transition occurs.

\section{Recursive multi-step predictive mixtures}
At horizon one, each simulation path begins from the observed final rolling window. At later horizons, each particle contains its own recursively simulated window, previous standardized state, and previous stabilized gate. For particle $b=1,\ldots,B$ and anchor $j$, \rc constructs
\begin{equation}
 Y_{h,b,j}=A_{h,b,j}+\gamma s_{h,b}\widetilde e_{h,b,j},
\end{equation}
with weight
\begin{equation}
 \omega_{h,b,j}=\frac{\widetilde\pi_{h,b,j}}{B}.
\end{equation}
The horizon-$h$ predictive distribution is therefore the explicit $BM$-point weighted mixture
\begin{equation}
 \widehat F_h(y)=\sum_{b=1}^{B}\sum_{j=1}^{M}\omega_{h,b,j}\mathbf{1}(Y_{h,b,j}\le y). \label{eq:empirical-cdf}
\end{equation}
Predictive means, variances, medians, and quantiles are computed from these weights before resampling.

To propagate paths, the implementation supports multinomial, systematic, and anchor-stratified systematic resampling. The default anchor-stratified scheme first allocates offspring counts approximately proportional to aggregate anchor mass, using integer floors plus largest remainders, and then selects particles within each anchor systematically. This reduces the chance that low-probability but non-negligible anchor components disappear from the recursive population solely because of Monte Carlo variation.

\section{Algorithm}

\begin{center}
\setlength{\fboxsep}{8pt}
\setlength{\fboxrule}{0.6pt}
\fbox{%
\begin{minipage}{0.94\linewidth}
\small
\textbf{Canonical Rollcast procedure}

\medskip
\textbf{Inputs.} Univariate levels $y_{1:T}$; rolling-window width $W$; anchor quantiles $\mathcal Q$; responsibility temperature $\tau$; gate penalty $\lambda$; nearest-state settings $(K,h_x)$; residual settings $(h_e,\gamma,\epsilon)$; persistence settings $(\rho_{\min},\rho_{\max},d_\rho)$; forecast horizon $H$; number of simulation paths $B$; resampling rule.

\medskip
\textbf{Outputs.} Horizon-specific predictive distributions and quantiles; simulated paths; raw and stabilized anchor probabilities; selected-anchor frequencies; adaptive-persistence diagnostics; fitted and tuning diagnostics.

\begin{enumerate}[leftmargin=2.2em,itemsep=0.45em,topsep=0.65em]
\item \textbf{Anchors and states.} For every observable forecast origin, compute the rolling anchor vector $A_t$, robust scale $s_t$, relative state $x_t$, standardized anchor residuals $e_t$, and soft anchor responsibilities $r_t$.

\item \textbf{Causal residual archive.} Standardize the state using training-period information. For each historical row and anchor, retrieve only earlier nearby states and combine state similarity with anchor responsibility to form a conditional residual distribution.

\item \textbf{Proper-score gate.} Fit the softmax gate $\pi_t$ by minimizing penalized negative log predictive density of the complete anchor mixture. When hyperparameters are searched, score every candidate on the same causal validation origins.

\item \textbf{Adaptive persistence.} Convert the raw gate into $\widetilde\pi_t$ using state-dependent inertia: stable states preserve more of the previous gate, while large state movements place more weight on the current gate.

\item \textbf{Recursive mixture.} At each horizon and for every particle, recompute anchors and state, draw a conditional residual for every anchor, and form all $M$ candidate next levels $A_{h,b,j}+\gamma s_{h,b}\widetilde e_{h,b,j}$ with weights $\widetilde\pi_{h,b,j}/B$.

\item \textbf{Forecast and propagate.} Aggregate the $BM$ candidates into the predictive distribution, compute requested summaries, resample parent--anchor candidates, update the particle windows and gate/state histories, and continue through horizon $H$.
\end{enumerate}

\vspace{0.2em}
\[
 y_{1:T}
 \;\longrightarrow\;
 (A_t,s_t,x_t,e_t,r_t)
 \;\longrightarrow\;
 \widehat f_{tj}
 \;\longrightarrow\;
 \pi_t
 \;\longrightarrow\;
 \widetilde\pi_t
 \;\longrightarrow\;
 \widehat F_{T+h}
 \;\longrightarrow\;
 \{Y^{(b)}_{T+1:T+H}\}_{b=1}^{B}.
\]
\end{minipage}}
\end{center}

\section{Practical statistical properties and validation design}
The relative-state formulation is insensitive to the measurement units of the series. If a series is shifted and multiplied by a positive constant, the default anchors and robust scale transform in the same way. As a result, the relative anchor positions, standardized residuals, responsibility weights, nearest-state geometry, and gate inputs remain unchanged; the predictive distribution simply shifts and rescales with the data. This is useful in practice because the behavior of the model is driven by local shape rather than by whether a series is expressed, for example, in units, thousands, or percentages. Exactly constant windows are handled by a small numerical scale floor.

The fitting and tuning design is also causal by construction. When a historical row is used to evaluate a component density, only earlier rows are eligible for nearest-state residual information. With an explicit validation block, state normalization and gate fitting use rows preceding the first validation origin. When alternative rolling windows are compared, the implementation chooses a common set of validation origins after the largest candidate window has accumulated sufficient history. This avoids giving different window lengths different realized target sets.

After hyperparameters have been selected, the deployment model is refitted on all valid causal-density rows with those values held fixed. In other words, validation is used for choosing the configuration, while the final gate recovers the available historical information before forecasting.

\section{Hyperparameter selection}
The public implementation follows a simple rule: a scalar hyperparameter is fixed; a vector denotes a candidate set to be searched. The anchor quantiles are structural definition parameters rather than tuning candidates. Candidate dimensions include $W$, $\tau$, $\lambda$, $K$, $h_x$, $h_e$, $\gamma$, responsibility smoothing $\epsilon$, and the persistence parameters $\rho_{\min}$, $\rho_{\max}$, and $d_\rho$.

A full Cartesian grid is usually unnecessary and can be expensive because causal nearest-state densities must be rebuilt for many configurations. \rc instead uses cached staged coordinate search on common validation origins. Starting from central values, parameters are visited in a fixed order and replaced only when a candidate lowers held-out negative log predictive density. One or more coordinate passes may be requested. With grid sizes $G_1,\ldots,G_p$, this changes the number of configuration evaluations from $\prod_k G_k$ for a Cartesian grid to approximately $\sum_k G_k$ per pass, although individual evaluations can still be computationally demanding.

The reference defaults are shown in Table~\ref{tab:defaults}. They are starting search spaces, not universal recommendations.

\begin{table}[t]
\centering
\caption{Reference search spaces in the supplied implementation. A scalar value fixes the parameter; a vector triggers staged search.}
\label{tab:defaults}
\small
\begin{tabular}{p{0.25\linewidth}p{0.61\linewidth}}
\toprule
Parameter & Default candidate values \\
\midrule
Rolling window $W$ & 20, 30, 45, 60, 90, 120 \\
Responsibility temperature $\tau$ & 0.15, 0.25, 0.40 \\
Gate penalty $\lambda$ & 0.001, 0.01, 0.05 \\
Nearest states $K$ & 20, 40, 80 \\
State bandwidth $h_x$ & 0.60, 1.00, 1.60 \\
Residual bandwidth $h_e$ & 0.20, 0.35, 0.55 \\
Residual strength $\gamma$ & 0, 0.10, 0.25, 0.50, 0.75, 1.00 \\
Responsibility smoothing $\epsilon$ & 0.01, 0.03, 0.06 \\
$\rho_{\min}$ & 0, 0.05, 0.10 \\
$\rho_{\max}$ & 0.80, 0.90, 0.97 \\
Persistence decay $d_\rho$ & 0.50, 1.00, 2.00 \\
Anchor quantiles & 0.05, 0.10, 0.25, 0.75, 0.90, 0.95 \\
Validation fraction & 0.25 \\
\bottomrule
\end{tabular}
\end{table}

\section{Relation to existing methods}
The closest literature is broader than forecast combination alone. Five strands are particularly relevant.

\paragraph{Forecast combination.} Bates and Granger established the basic case for combining forecasts, while later reviews documented the robustness of simple combinations and the expansion toward adaptive, nonlinear, and probabilistic schemes \citep{BatesGranger1969,Clemen1989,WangEtAl2023}. \rc differs in the objects being combined: statistical functionals of one rolling empirical window rather than forecasts from separately fitted models.

\paragraph{Predictive-density pooling and proper scores.} Hall and Mitchell estimate density-combination weights using a Kullback--Leibler/log-score criterion, and Geweke and Amisano show why optimal linear prediction pools can retain several components with positive weight even when individual models differ in quality \citep{HallMitchell2007,GewekeAmisano2011}. Proper scoring rules provide the decision-theoretic basis for evaluating probabilistic forecasts \citep{GneitingRaftery2007}. In \rc, mixture weights are conditional on anchor-derived state variables and the component densities come from local residual archives.

\paragraph{Local analog and nearest-state forecasting.} Local prediction based on similar past states has a long history in nonlinear time-series forecasting \citep{FarmerSidorowich1987,SugiharaMay1990}. \rc shares the idea that past neighborhoods can be more informative than a single global equation, but it does not forecast directly from the future trajectories of nearest neighbors. Instead, neighbor information is used to estimate residual distributions around each rolling anchor.

\paragraph{Conditional-density estimation.} Kernel and local-polynomial approaches estimate the distribution of a response conditional on predictors rather than only its conditional mean \citep{HyndmanEtAl1996,HyndmanYao2002}. The residual layer in \rc belongs conceptually to this family, although it is intentionally simpler: the conditioning variables are the standardized anchor state, the archive is restricted causally, and neighbor weights are multiplied by anchor responsibilities before kernel jitter is applied.

\paragraph{Dynamic and feature-dependent mixtures.} Dynamic model averaging allows model probabilities to evolve through time, feature-based density combinations let weights depend on observed time-varying features, and mixture-of-experts models learn routers that specialize experts across patterns \citep{RafteryEtAl2010,LiKangLi2023,JacobsEtAl1991,NiEtAl2024}. \rc uses a much lower-capacity version of this idea. Its gate is linear in standardized anchor geometry, and persistence is a deterministic function of observed state movement rather than a latent transition model.

\section{Synthetic calibration benchmark}
\subsection{Design and benchmark}
The CRAN implementation of \texttt{rollcast} version 0.1.0 was evaluated on eight fixed data-generating processes (DGPs) covering stationary, integrated, trending, nonlinear, regime-switching, heteroskedastic, heavy-tailed, and variance-break dynamics. Each DGP contributes 250 independently generated replications with 300 training observations and a six-step realized future path, for 2,000 fitted series and 12,000 forecast targets. Rollcast uses 1,000 particles with anchor-stratified resampling.

\begin{table}[t]
\centering
\caption{Synthetic data-generating processes. Innovations are Gaussian unless otherwise stated.}
\label{tab:dgps}
\small
\begin{tabular}{p{0.25\linewidth}p{0.64\linewidth}}
\toprule
DGP & Mechanism \\
\midrule
Gaussian AR(1) & \(y_t=0.70y_{t-1}+\varepsilon_t\). \\
Gaussian random walk & \(y_t=y_{t-1}+\varepsilon_t\). \\
Local trend & \(d_t=0.85d_{t-1}+0.15\eta_t\), \quad \(y_t=y_{t-1}+d_t+0.50\varepsilon_t\). \\
Threshold AR & \(y_t=\phi(y_{t-1})y_{t-1}+\varepsilon_t\), with \(\phi=0.85\) above zero and \(0.20\) otherwise. \\
Markov switching & Two regimes with \((\mu_s,\phi_s,\sigma_s)=(-1.5,0.30,0.60)\) and \((1.5,0.80,1.50)\), with transition matrix \(\left(\begin{smallmatrix}0.95&0.05\\0.08&0.92\end{smallmatrix}\right)\). \\
Stochastic volatility & \(h_t=-0.20+0.95(h_{t-1}+0.20)+0.20\eta_t\), \quad \(y_t=0.50y_{t-1}+\exp(h_t/2)\varepsilon_t\). \\
Heavy-tail AR & \(y_t=0.60y_{t-1}+u_t\), with variance-standardized Student-\(t_5\) innovations. \\
Variance break & \(y_t=0.50y_{t-1}+\sigma_t\varepsilon_t\); \(\sigma_t\) rises from \(0.60\) to \(2.00\) for the final 25 training observations and remains high over the forecast horizon. \\
\bottomrule
\end{tabular}
\end{table}

The calibration-related search varies \(W\in\{30,60,90\}\), \(h_e\in\{0.20,0.35,0.55\}\), residual strength \(\gamma\in\{0.10,0.25,0.50,0.75,1.00\}\), and persistence decay \(d_\rho\in\{0.50,1.00,2.00\}\); the remaining structural settings are fixed. Candidate configurations are selected by causal proper-log-score validation on common origins and refitted on the full training history.

For every replication, 10,000 paths from the known DGP conditional on its terminal observed and latent state define an approximate oracle predictive distribution. The experiment therefore measures calibration and probabilistic efficiency relative to the true conditional law. It is not a comparison against practical fitted alternatives such as ARIMA, ETS, or GARCH. Coverage is evaluated at 50\%, 80\%, 90\%, and 95\%; the results below emphasize the 90\% and 95\% intervals. For each fixed DGP--horizon cell, the 250 replications are independent, and Wilson intervals and exact binomial diagnostics are computed at the cell level \citep{Wilson1927}. CRPS, WIS, interval width, and interval score are also calculated; oracle ratios equal one when Rollcast matches the oracle benchmark.

\subsection{Results}
Table~\ref{tab:dgp-performance} gives the main comparison with the oracle. Across all DGP--horizon cells, mean coverage is 0.862 for nominal 90\% intervals and 0.915 for nominal 95\% intervals, compared with oracle coverages of 0.901 and 0.952. Rollcast intervals are 13.6\% wider than the oracle at the 90\% level and 15.9\% wider at the 95\% level. Overall normalized CRPS is 14.4\% above the oracle. Wider intervals therefore do not remove the coverage gap; conditional location and distributional-shape errors remain relevant.

\begin{table}[t]
\centering
\caption{Rollcast calibration and probabilistic efficiency relative to the true-DGP oracle, averaged over horizons one through six. CRPS/oracle and width/oracle equal one at oracle performance.}
\label{tab:dgp-performance}
\small
\begin{tabular}{lrrrr}
\toprule
DGP & 90\% cov. & 95\% cov. & 90\% width/oracle & CRPS/oracle \\
\midrule
Gaussian AR(1) & 0.868 & 0.922 & 0.999 & 1.098 \\
Gaussian random walk & 0.877 & 0.933 & 1.284 & 1.196 \\
Local trend & 0.865 & 0.917 & 1.659 & 1.461 \\
Threshold AR & 0.845 & 0.908 & 1.002 & 1.100 \\
Markov switching & 0.861 & 0.900 & 1.098 & 1.287 \\
Stochastic volatility & 0.864 & 0.921 & 1.004 & 1.081 \\
Heavy-tail AR & 0.859 & 0.911 & 1.027 & 1.088 \\
Variance break & 0.855 & 0.911 & 1.012 & 1.109 \\
\midrule
Overall & 0.862 & 0.915 & 1.136 & 1.144 \\
\bottomrule
\end{tabular}
\end{table}

Calibration weakens with forecast horizon. Mean 90\% coverage falls from 0.898 at \(h=1\) to 0.840 at \(h=6\); mean 95\% coverage falls from 0.940 to 0.901. Local trend and Markov switching show the largest CRPS losses relative to the oracle, at 1.461 and 1.287 respectively. The other six DGPs range from 1.081 to 1.196, with the smallest gap under stochastic volatility.

The adaptive search reacts strongly to the variance break: \(\gamma=1\) is selected in 99.2\% of fits, \(h_e=0.55\) in 95.2\%, and \(W=30\) in 54.8\%. All gate optimizations converge. The persistence-stabilized training negative log score is lower than the raw-gate score in only 0.65\% of fits, with a mean stabilized-minus-raw difference of \(+0.035\); persistence therefore acts primarily as a smoothing regularizer in this experiment.

\section{Computational considerations}
The method is intentionally lightweight in model form but not necessarily in brute-force training cost. For each forecast origin, anchors require rolling statistics and quantiles. Causal conditional density construction then performs nearest-state search against an expanding historical archive. Without a dedicated neighbor index, this stage is approximately quadratic in the number of training states for a single full density build. The Rcpp implementation mitigates constants by maintaining only the $K$ nearest candidates in a heap rather than sorting every distance.

Recursive forecasting is particle based. At each horizon, $B$ particles each generate $M$ anchor candidates, producing a $BM$-point weighted mixture before resampling. The main cost comes from recomputing rolling anchors and nearest-state residual retrieval for each particle. The current implementation is suited to moderate-dimensional univariate forecasting rather than high-throughput global forecasting. Approximate nearest-neighbor indexing, cached rolling quantiles, and parallel particle updates would reduce computational cost.

\section{Discussion}
\rc connects forecast combination, conditional-density estimation, nearest-state prediction, and mixture-of-experts methods through a compact anchor representation. Unlike conventional forecast pools, the components are rolling statistical functionals rather than separately fitted forecasting models. Unlike direct analog forecasting, nearest historical states are used to estimate anchor-specific residual distributions. The gate then assigns state-dependent probability to the resulting component densities using a proper log score.

The synthetic benchmark identifies a consistent calibration gap. Nominal 90\% and 95\% intervals cover 86.2\% and 91.5\% of outcomes on average, and the gap increases with horizon. Because the corresponding intervals are wider than the oracle, the dominant error is not simple under-dispersion. The recursive distribution can be wide and still place probability mass in the wrong location or shape.

The DGP results separate different failure modes. Local trend produces intervals 65.9\% wider than the oracle and a CRPS ratio of 1.461, showing poor dynamic tracking despite generous dispersion. Markov switching produces a CRPS ratio of 1.287, consistent with latent regimes that are only indirectly represented by anchor geometry. Stochastic volatility and heavy-tailed autoregression are closer to the oracle, with CRPS ratios of 1.081 and 1.088, indicating that the conditional residual archive can represent changing dispersion and heavy tails when comparable observed states recur.

The simulation benchmark has clear limits. The eight DGPs are fixed stress scenarios, so cross-DGP averages are descriptive. The oracle uses the true model and latent state and is therefore an efficiency reference, not a realistic fitted competitor. The experiment does not establish comparative superiority over ARIMA, ETS, GARCH, state-space, or machine-learning alternatives. The current anchor dictionary also omits explicit seasonal, long-memory, calendar, and multivariate structure, while recursive simulation compounds one-step approximation errors. Finally, persistence rarely improves the training log score in this experiment, which motivates a direct persistence ablation.

The observed horizon-dependent undercoverage motivates horizon-aware residual calibration or horizon-dependent residual strength. The local-trend and Markov-switching results motivate richer state features or specialized anchors, while CRPS- or interval-score-based tuning would align parameter selection more directly with distributional calibration \citep{GneitingRanjan2011}.

\section{Conclusion}
\rc is an adaptive probabilistic forecasting framework that treats rolling statistical summaries as forecast anchors and combines them through state-dependent proper-score gating, causal residual conditioning, adaptive persistence, and recursive all-anchor resampling.

In the synthetic benchmark, the CRAN implementation completes all 2,000 fits and remains relatively close to the true-DGP oracle for several stationary and heteroskedastic mechanisms. Mean coverage is 0.862 for nominal 90\% intervals and 0.915 for nominal 95\% intervals, while those intervals are wider than the oracle on average. Calibration deteriorates with forecast horizon, and local trend and Markov-switching dynamics produce the largest efficiency losses.

The results identify multi-step conditional calibration as the main limitation of the current implementation. Horizon-aware calibration, persistence ablation, richer state representations, and anchors specialized for trend or regime dynamics are the most direct extensions suggested by the benchmark.

\section*{Software and reproducibility}
The reference implementation is the R package \texttt{rollcast}, with Rcpp-compiled routines for causal component-density construction and recursive simulation. The package can be installed directly from CRAN with \texttt{install.packages("rollcast")}. The package page and source archive are available at:
\begin{center}
\url{https://cran.r-project.org/web/packages/rollcast/index.html}
\end{center}
The reported benchmark was generated with R 4.5.1 and the CRAN \texttt{rollcast} 0.1.0 package. The main protocol uses eight fixed DGPs, 250 independent replications per DGP, 300 training observations, horizons one through six, 1,000 Rollcast particles, and 10,000 oracle trajectories per replication. The protocol script, raw forecast and interval rows, diagnostics, metadata, and summary files are included in the accompanying reproducibility bundle.

\end{document}